\documentclass[10pt, conference, compsocconf]{IEEEtran}
\usepackage{graphicx}
\usepackage{enumitem}
\usepackage{array}
\usepackage{subcaption}
\usepackage{ltablex} 
\usepackage{hyperref}       
\usepackage{url}
\usepackage{supertabular,booktabs}

\begin{document}

\title{Making Smart Homes Smarter: Optimizing Energy Consumption with Human in the Loop}

\author{\IEEEauthorblockN{Mudit Verma}
\IEEEauthorblockA{\textit{Delhi Technological University} \\
New Delhi, India \\
mudit.verma2014@gmail.com}

\and

\IEEEauthorblockN{Siddhant Bhambri}
\IEEEauthorblockA{\textit{Delhi Technological University} \\
New Delhi, India \\
siddhant4dtu@gmail.com}

\and

\IEEEauthorblockN{Saurabh Gupta}
\IEEEauthorblockA{\textit{IIIT-Delhi} \\
New Delhi, India \\
saurabhg@iiitd.ac.in}

\and

\IEEEauthorblockN{Arun Balaji Buduru}
\IEEEauthorblockA{\textit{IIIT-Delhi} \\
New Delhi, India \\
arunb@iiitd.ac.in}
}

\maketitle

\begin{abstract}
Rapid advancements in the Internet of Things (IoT) have facilitated more efficient deployment of smart environment solutions for specific user requirement. With the increase in the number of IoT devices, it has become difficult for the user to control or operate every individual smart device into achieving some desired goal like optimized power consumption, scheduled appliance running time, etc. Furthermore, existing solutions to automatically adapt the IoT devices are not capable enough to incorporate the user behaviour. This paper presents a novel approach to accurately configure IoT devices while achieving the twin objectives of energy optimization along with conforming to user preferences. Our work comprises of unsupervised clustering of devices’ data to find the states of operation for each device, followed by probabilistically analysing user behavior to determine their preferred states. Eventually, we deploy an online reinforcement learning (RL) agent to find the best device settings automatically. Results for three different smart homes’ data-sets show the effectiveness of our methodology. To the best of our knowledge, this is the first time that a practical approach has been adopted to achieve the above mentioned objectives without any human interaction within the system. 
\end{abstract}

\begin{IEEEkeywords}
Internet of Things; Unsupervised Clustering; Markov Decision Processes; Reinforcement Learning 
\end{IEEEkeywords}

\section{Introduction}
\label{intro}
The increasing involvement of smart devices in our lives has necessitated us to come up with better coordination and management strategies for increased efficiencies and Quality of Service (QoS). This increase in the number of devices has resulted in higher energy consumption by households and has led to the problem of energy optimization. 
Many organizations are investing in extensive research to come up with strategies to make utilization of energy as efficient as possible. “Smart Homes” or “Smart Devices” is one such area that makes use of Information \& Communication Technologies for finding solutions to such environmental issues. With the help of embedded intelligence, such devices can target the preferences of a user during everyday life. However, cogency of such results is critical \cite{8,9}. With the installation of smart meters along with these devices at homes, we can find effective ways to provide users with a visualization of their energy consumption pattern in a way they can comprehend \cite{5}. Furthermore, we can even find ways to make intelligent systems that can recommend users or act on their behalf, based on their past power consumption behavior, to achieve the goal of minimizing electricity wastage. The objective here needs to be to prevent any unnecessary consumption of energy, preserving the comfort and productivity of the user.
In smart homes, a vast number of heterogeneous appliances, sensors, and actuators inter-operate and provide context information, which in turn, together with user preferences, are used to effectuate a value-added functionality dynamically. A smart home needs to be able to analyse the actions of its occupant, taking into account the context information, to proactively recognize the occupant’s activity to conserve energy. Therefore, while creating such systems, one must proceed in a way such that the intelligent planner’s policies in promoting efficient energy usage in households must take into account the user’s choices and behavior for utmost user satisfaction \cite{7}. 

Another study of devices reveals a framework where cluster analysis is done in a three-part architecture, starting with maintaining the sequence of individual’s activities during the day, showing how long each activity is carried out and ending with finding the clusters for different activity patterns for the group of users \cite{1}. However, the results from this analysis have not been put to any efficient use for the user. Another intelligent planning approach has three entities: the manufacturers who provide the state information, the SESPs (Smart Environment Service Providers) who manually assess the device states and the users get a recommended set of actions to reduce energy consumption \cite{2}. The two frameworks combined cater to the problem, but a lot of manual effort is involved and therefore, faces a lot of challenges regarding the feasibility. Hence, we aim to overcome these challenges by applying a very simple yet effective approach to automate the entire framework from collecting and analysing user data to taking decisions in accordance with the user’s preferences without involving any manual effort of any kind, once the framework is deployed.

We first elaborate on these and some other challenges along with solutions focused on only some specific parts of the complete problem at hand in Section 2. In Section 3, we explain our approach to adaptively configure smart devices in a smart home through the effective incorporation of probabilistic user behaviour(s). Then, we present a method to automate the complete process of collecting the state information (i.e., finding clusters for different activity patterns), followed by analysing it to define the quality of states (i.e., assess the device states keeping in mind the user preferences). Eventually, we formulate a RL technique for recommending actions to the user to achieve the set goal of optimal power consumption in Section 4. We conclude by presenting the results of our approach in Section 5 followed by the conclusion in Section 6.
 
\section{Related Work}
\label{background}
The increase in computing power in constrained environments have allowed researchers to experiment and come up with new and better approaches to solve some critical problems using the power of artificial intelligence. Hence, there have been some approaches proposed earlier which either have been mainly theoretical in nature or have focused only on one specific part of this problem statement. We present a brief discussion on these methodologies adopted by researchers in the past.

In works that study the behavioural clustering of devices, Adika et.al. present a method where appliances are clustered together based on their hourly energy consumption data \cite{12}. Then a “time of use” probability distribution is made, and each cluster is given a schedule, making this a job scheduling problem which is solved using dynamic programming. The challenge here is to predict the correct cost of energy for the next hour. Consequently, the scheduling assumes that the error in predicting the cost is minimum. However, the approach does not focus on any method to suggest the user ways to cut down on excessive power consumption.

In yet another approach, the authors introduce a generic way of creating a Peer to Peer (P2P) overlay network using Hydra middle ware \cite{6, 10}. The authors talk about collecting and displaying energy consumption data using several data visualisation techniques. The approach does not have intelligent monitoring or recommendation for the users, which makes us ask the question of whether the data visualisation effects the user’s understanding of energy consumption or not \cite{5}. 

Wei et. al. show another general architecture for designing the energy consumption monitoring and energy-saving management system that are IoT based \cite{11}. According to this work, IoT infrastructure has three levels: the bottom-most level collects data from different sensors, the middle level which is the network layer talks about data transmission, and the topmost level processes the collected and transferred data using cloud computing and fuzzy pattern recognition techniques. A 3-component architecture was also introduced to cater to an intelligent household lighting system for efficient energy consumption, including user’s context information such that the system’s behaviour conforms to user satisfaction \cite{13}. Although this approach talks about an intelligent planner, it is limited to the lighting systems and uses a static algorithm (minimum light intensity control) to control the lights. Hence, the approach is not efficient when it comes to dynamic environments where user behaviour is prone to change or in a real-life setting where a set of diverse electrical equipment or appliances are present.

Yau et. al. base their technique on Markov Decision Process (MDP), assuming there are three entities \cite{2}. Firstly, the manufacturer provides the set of device states, secondly a SESP (Smart Environment Service Provider) who analyses the state information and lastly, the MDP planning algorithm which makes a policy of states and actions along with a user who receives these actions suggested by the MDP and acts accordingly in the given environment. The approach assumes a Central Module which collects all the information and does the heavy lifting, in turn, solving the problem of having a constrained environment. The approach involves manual human intervention, firstly as the manufacturer who supplies information on the devices and then in the role of SESP who performs analysis on the data. Also, The results are shown over simulation, and the approach is never used on a real data-set. 

The methodology used by Jahn et. al. and Byun et. al. makes us understand the major approaches for collecting data from appliances and hence, helps us move in the direction of intelligent planning \cite{6, 13}. The architecture proposed by Wei et. al. is a general way to move from the stage of data collection to intelligent planning \cite{11}. There are existing approaches that are perfectly capable of achieving optimal energy consumption, taking the assumption that the user behaviour is not dynamic, and the price prediction has minimum error \cite{12, 13}. The approach used by Yau et. al. uses RL and hence, is capable of dealing with a change in user behaviour \cite{2}. However, the role of entities like manufacturer and SESPs can be automated. In this work, we find the quality of states based on the user behaviour, which keeps on updating itself to improve with time. The work presented in this paper is an advancement of the approach discussed in \cite{2}.

\section{Proposed Methodology}
\label{methodology}

We use a top-down strategy to explain the complete method. First, we give an overview of the whole system and interactions between logical sub-sections. Then we dissect each part and discuss our approach in depth.

\subsection{Flow of Data}
\label{dataflow}

\begin{figure}[htbp]
\centering
\includegraphics[width=0.45\textwidth]{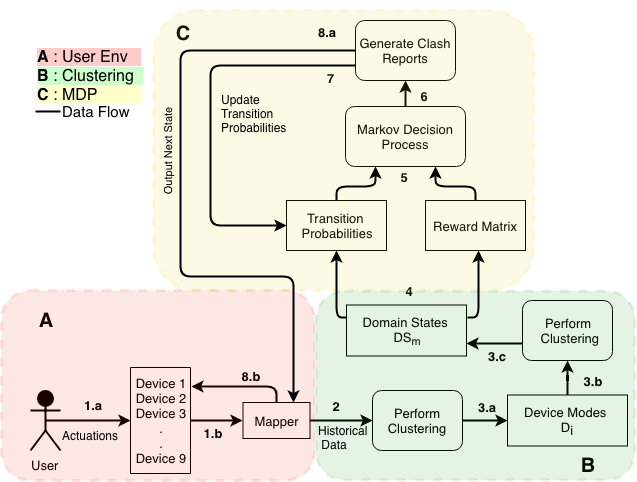}
\caption{The flow of data in proposed framework.} 
\label{architecture}
\end{figure}

According to Figure \ref{architecture}, the model is divided into three segments. 

\textbf{Segment A}: This is the first segment in the entire flow of data inside our model. The data is collected from the user, and his/her behaviour/preferences are recorded, which comprise of an \textit{Action} and an \textit{Actuation} as shown in steps 1.a and 1.b of the figure. We define the actual actions taken by the user to change the device settings as \textit{Actuations} whereas the stochastic actions relevant to the RL agent are termed as \textit{Actions}, and these are received by the \textit{Mapper}.

\textbf{Segment B}:  In the second segment, the devices' power consumption values are clustered to determine device modes, ${D_i}^{n_i}$ for $i^{th}$ device, where the device has $n_i$ modes of operation. These device modes are again clustered together to form $m$ \textit{domain states} ${DS} = \{ {DS_1},{DS_2}...{DS_m} \}$. Hence, $ DS_i = ({D_1}^{n_1}, {D_2}^{n_2}, ... {D_k}^{n_k}) $ for k devices. Therefore, $D_i$ signifies device-level representation whereas $DS_i$ is representative of the entire set of devices. 

\textbf{Segment C}: The domain states act as an input to the third and last section, i.e. Segment C, where our RL algorithm takes a suitable decision with respect to choosing a domain state optimally. \textit{Transition Probability Matrix} and \textit{Reward Matrix} are two essential components of our RL algorithm which are fed as inputs and these are computed using the Domain States. The RL agent generates an output and compares these with the historical data during the learning phase. It suitably adapts to the user’s behavior and updates the probabilities of transitioning from one domain state to another.

The \textbf{Mapper} entity in segment A of Figure \ref{architecture} is simulated for the results in our paper as mentioned in sections \ref{actionsactuations} and \ref{validation}. It is a many-to-many mapping where several actual device settings can correspond to internal device modes and vice versa. This mapping simplifies our approach to search for the appropriate device modes which a user can toggle through a suitable actuation. The live data input source will require recording the actuation signalled by the user and the corresponding device mode action. This mapping may also be clubbed with the clustering algorithms as a “feature input” so that the clustered nodes also have information about the actuation. Other methods may involve finding the best possible actuation (such as by using costs incurred in shifting device modes) which corresponds to some device action in step 8.b of Figure \ref{architecture}.

\subsection{Two-Step Clustering}
\label{two step clustering}
As our first step in our proposed approach, we intend to remove the use of SESPs \cite{2} who manually provide device modes \& label them as good or bad. This manual labour inhibits the system to achieve complete automation. The two-step clustering refers to using the available raw user data to form device mode clusters followed by using these device modes to form domains state clusters.

We use the data of user actions and perform clustering using time features (hour, month \& year) and device power consumption for device modes, where a device is not aware of the existence of another device which ensures independence of device usage, a qualitative feature of appliances/devices in real-life. Hence, device modes are clustered output equivalents of device actuations as discussed in \ref{dataflow}. Domain states, on the other hand, are clustered outputs which take all the devices into consideration, wherein the states are built up by the use of all of the devices’ modes clustered together. 

We use Growing Neural Gas clustering algorithm which yields us with a graph when the input data is projected onto a 2D space. Each vertex of this graph corresponds to a neuron in which input data has been mapped. GNG learns through a combination of updated Kohonen learning approach to adjust the positions of the neurons along with a Competitive Hebbian Learning (CHL) approach for its connections \cite{gng}. The number of connected components in this graph represent the total number of clusters.

\subsection{Actions and Actuations}
\label{actionsactuations}
As defined in Segment A of \ref{dataflow}, we further explain these two terms. Initially, we consider a user’s historical data to supply the actions, i.e., if the current domain state and previous domain state are same, it is taken to be a \textit{STAY} action, otherwise, \textit{MOVE}. For example, let the current domain state for 5 devices be (1,1,0,3,2) and the next domain state for the same 5 devices be either (1,1,0,3,2) or (1,1,0,2,0). In the first scenario, where the device mode values remain the same for all 5 devices, it is said to be a \textit{STAY} action whereas in the second case wherein the device mode values changed for the last two devices, it is said to be a \textit{MOVE} action.

To obtain the actuations, we select 30\% of our training data points for which actions differ from the actuations. This implies that the actions and actuations are drawn from a “similar” distribution. Thus, there are certain data points where a \textit{STAY} action may have resulted in \textit{MOVE} and vice-versa. Hence, these 30\% positions signify the underlying stochasticity in the model. We label the time-stamps signifying the actual \textit{MOVE} and \textit{STAY} actions as $M$ and $S$, and the states representing actuations as $m$ and $s$.

\begin{figure}
\centering
\includegraphics[width=0.45\textwidth]{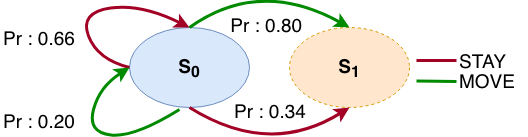}
\caption{Possibilities available to a state given the actions: STAY and MOVE} 
\label{stateDiag}
\end{figure}

Now, the model has a probability $P_{dd'}(i \cap j)$ which represents the state transition from domain state $d$ to $d′$ when actuation $i$ and action $j$ occurred, where $i \in \{m,s\}$ and $j \in \{M,S\}$. However, these actuations are for training dataset only. For testing data, we find actuations as in section \ref{validation}.
Figure \ref{stateDiag} shows a sample transition between any two random domain states $s_0$ and $s_1$ with the agent currently at $s_0$. Due to the attached stochasticity, both \textit{MOVE} and \textit{STAY} action can result in a transition from $s_0$ to $s_1$ or transition back to the same state $s_0$. Note that the sum of probabilities for all transitions as \textit{MOVE} or \textit{STAY} is unity, which is implicit.

\subsection{Domains State Types}
For understanding the domain states qualitatively, we examined the training data set and divided these domain states into two broad categories- high demand (HD) states and low demand (LD) states. The HD states are the most frequently visited states by the user whereas, on the other hand, LD states are the least visited states. Let us say that top percent of the total number of domain states comprise of HD states and the remaining are the LD states.

In a real-life scenario, there may be some domain states which the user may prefer to stick to, irrespective of the output of the RL agent and these domain states could intuitively fall under both the categories. We refer such domain states as strict high demand (SHD) states and say, that the HD states comprise of $fix_{HD}$ percent of SHD states and the remaining HD states are termed as loose high demand (LHD) states for simple reference. Similarly, there may be some infrequently occurring states, i.e. LHD states which the user may prefer to stick to. We refer such domain states as strict low demand (SLD) states and say, that the LD states comprise of $fix_{LD}$ percent of SLD states and the remaining LD states are termed as loose low demand (LLD) states for simple reference.

\subsection{Reinforcement Learning Agent}

For our RL agent formulation, we intended to experiment with a simple RL model and hence used the MDP algorithm which provided us satisfactory results. Therefore, this section highlights our MDP formulation for the given problem setting. Henceforth, this provides us with a future opportunity of modelling the current problem statement using more complex RL algorithms and compare the results as shown in the current approach. 
A Markov Decision Process is a reinterpretation of Markov Chains which includes an agent and a decision-making process \cite{17}. Our MDP is characterised by the following components: 

\begin{enumerate}
\item State Space: $S={d_0,d_1,...,d_m}$ where $d_i$ represents $i^{th}$ domain state,
\item Initial State: $d_0$ the first domain state in data-set,
\item Actions: $A={a_0,a_1}$ where $a_0$ is STAY and $a_1$ is MOVE,
\item Transition Model: $T(s,a,s')$ and 
\item Reward Function: $R(s)$ = $-1 * totalPowerConsumptionOfDomainState$. 

\end{enumerate}

\subsubsection{State Space}
The clustering of domain states provides an implicit functionality of state space reduction for MDP; therefore, preventing state space explosion. It is possible to perform domain state clustering directly, without the use of first level device mode calculation. However, to enforce the MDP outputs on the system, we must know to which device mode MDP is asking the devices to move, and thus the need for finding device modes.

\subsubsection{Transition Matrix}
The transition matrix returns the probability of reaching the state $s’$ if action $a$ is taken in state $s$. We use the training data set to compute this matrix. These transitions are governed by the reward received by the RL agent on taking a particular action $a$ in state $s$ to reach a new state $s’$. The Reward function $R(s)$ returns a real value every time the agent moves from one state to the other. The agent receives a higher reward (less negative in value) when it reaches one of these desirable states. 

\subsubsection{Updating Transition Matrix}
The RL agent gets the current domain state as the input and finds the state the user should be in, which is the output of the MDP function $mdp(s_t)$. While learning on the training data, the MDP agent calculates the number of clashes which take place when the user transitioned to another state while the RL agent recommended a different state. Our MDP algorithm runs online and is, therefore, able to adapt during the testing phase as well. When such a clash occurs, an update of the Transition Matrix involves reducing the probabilities of MDP outputs by a factor $e=0.1$ and increasing the probabilities of states belonging to the sets SHD and SLD, finally normalizing to maintain the probability sum as unity.

\subsubsection{MDP Algorithm}

The user actions have been modelled via the transition matrix, whereas the need for optimal power consumption is taken care of by the Reward Function. We implement our MDP using policy iteration algorithm where the agent chooses the best state using a policy. Policy iteration is guaranteed to converge to the optimal policy, and it often takes fewer iterations to converge than value iteration. The maximum expected utility principle states that a rational agent should choose its action that maximizes its maximum utility \cite{15}. The utility of a single state is defined as:
 
\begin{equation}
U(s) = E[\sum_{t=0}^{\infty} \gamma^{t}R(s_{t})] 
\end{equation}

The utility of a state $s$ is correlated with the utility of its neighbor at $s'$ as:

\begin{equation}
U(s) = R(s) + \gamma max\sum_{s'}T(s,a,s')U(s') \;\;\;;\; \; \gamma=0.9
\end{equation}

\subsubsection{Reward Function}
To account for the objective of reducing excessive power consumption, we planned to make use of a simple reward function that caters to our aim of rewarding the RL agent to make a transition to a state where the total power consumption is less than that at the current state. Hence, we want to penalize the domain states with higher power consumption and reward the domain states with lower power consumption. Therefore, we negate the parameter $totalPowerConsumptionOfDomainState$ to imply a more substantial reward for the domain states with lower power consumption.

\subsubsection{Reward Matrix}
The reward matrix is calculated by first finding the domain states, each of which has devices operating in specific modes, thus consuming power. We sum the consumption values to get a reward value for each domain state. The reward matrix is formed using these calculated reward values for all states. The current reward function penalizes a state for using power. The reward matrix can be created dynamically since with our clustering approach; we get a set of connected components of nodes rather than one single cluster centre. Therefore, a single device mode corresponds to the connected component of the graph. For our results, we use the average power consumption value for all the nodes in a component. However, we may also choose different power consumption values for different nodes to represent the power consumption of the device mode and have a more ``intelligen'' reward matrix.

\subsection{Method of Validation}
Comparing methods for their efficacy in energy optimality is a reasonably easy task by comparing the expected energy consumed. However, this is not so trivial for proving that our approach would work well at preserving human device setting preference. There is no supervised information present as to which states are liked or disliked by the user. Neither can a state lasting over a long time be considered as a desired state nor can an impulsive change from one state to another be labelled as a disliked state. Since application functionality of the devices is not studied; to centrally prevent the need of manual domain experts, we cannot find more likely or less likely preferred states usage-wise solely by the power consumption values. Thus, we require to prove two statements true simultaneously regarding our system to justify the aim of this paper effectively:

\begin{enumerate}
\item Reduction in clash rate over time, primarily for HD states.
\item Decrease in overall power consumption.
\end{enumerate}

We define Clash Rate as the number of incorrect next state predictions made by the MDP for the actual future (next time step from the testing data) per time-slot. It is given by:

\begin{equation}
\begin{array}{l}
{s'}_{t+1} = mdp(s_t) \;\;\; and \;\;\;  s'_{t+1} \ne s_{t+1} \;\;\;\; \\
s_{t+1} \in set(sHD+randomLHD) 
\end{array}
\end{equation}

Here, subscripts are time steps; $s'_{t+1}$ is obtained via the function $mdp(s_t)$ which takes the previous state $s_t$ as input. Thus,  $s'_{t+1}$ does not belong to the set of SHD or SLD states. 

Reduction in such clashes indicates that MDP can learn user state preferences, even if they are costly in terms of power consumption. Now, clash rate reduction should preferably occur for these states, since we need to preserve what the user likes and let the MDP work on the remaining dispensable states which can help us to reduce the power consumption. 

The first requirement shows that MDP understands the user preferences, and further, we can justify that our system achieves its primary objective of optimizing power consumption along with user comfort. We show that this power consumption efficiency comes from swapping most SLD states with the state outputs of the MDP.

\section{Experiments}

All experiments have been performed on Ubuntu-16.04, Intel i5 Processor, and 16GB memory, which indicates that the proposed model can be deployed on personal computers efficiently.

\subsection{Data-Set Description}
For this study, we apply a real-life data-set that is part of the UMass Trace Repository and taken from Smart* Data-set for Sustainability\footnote{http://traces.cs.umass.edu/index.php/Smart/Smart}. Barker et. al. present a way to collect long-term data, which makes it easier and efficient to map a user’s behaviour \cite{4}. The data contains the average electricity usage of several smart homes collected over regular intervals of 15 seconds for the following devices: Refrigerator, Freezer, Washer Dryer, Washing Machine, Dishwasher, Computer, Television, and Electric Heater. These devices have smart meters attached to them to record the power consumption after regular time intervals. We have chosen the data for smart homes A, B, and C. The data-set collected here is focused on sensing depth, i.e., collecting as much data as possible from each home, rather than breadth, i.e., collecting data from as many homes as possible.

Further, the data-set is divided into two parts: train and test. We use the training data for calculating MDP parameters, and the testing data to simulate live user interaction with our system. There is no distinction between training and testing data other than the fact that the training data is historical. Therefore, we assume it to be available from the beginning of the MDP run. Hence, all pre-processing and analysis are done on it. On the other hand, the testing data is present as live data with which we try to simulate the actual interaction of MDP with the user.

\subsection{Domain State Clustering}

\begin{table}[htbp]
\caption{Hyper-parameters for Growing Neural Gas Clustering.}
\label{gngparams}
\centering
\begin{tabular}{|l|l|l|}
\hline
\textbf{GNG Parameters}                                                           & \textbf{Device Mode} & \textbf{Domain State}\\ \hline
Max No. of Nodes                                                          & 10000       & 20000        \\ \hline
Max. Edge Age                                                             & 100         & 50           \\ \hline
Decay Rate: After Split        & 0.5         & 0.3          \\ \hline
Decay Rate: Error               & 0.995       & 0.9          \\ \hline
\end{tabular}
\end{table}

We considered clustering experiments using different algorithms such as DBSCAN, K-Means, and Growing Neural Gas (GNG). S. Dang explains the disadvantages of DBSCAN along with other clustering algorithms and states that density-based algorithms like DBSCAN do not take into account the topological structuring of the data, which is well mapped by the graphical modelling that GNG performs \cite{3}. It is because of GNG’s incremental nature that it is not necessary to decide on the number of nodes to use a priori, unlike the k-means clustering algorithm, where several trials may be required to determine an appropriate number of centres.
GNG is, therefore, suitable for problems where we know nothing or little about the input distribution, or the cases in which deciding on network size and decaying parameters are very difficult or impossible. Hence, we base our results on GNG being the primary clustering algorithm.

Bernd explains the use of GNG in its entirety \cite{16}. However, in our experiments, GNG had to be changed at certain steps where the algorithm is not explicit; for example, tackling the situation when a node does not have a neighbour and experimenting with the distance metrics used. We obtain device mode/state graphs and further find connected components that represent different data clusters. The algorithm is accurately able to find the number of classes and the associated neurons. Our prediction is based on the K-Nearest Neighbour approach upon GNG nodes, where we set k = 3. The hyper-parameters for clustering have been shown in \ref{gngparams}. The general parameters such as number of epochs is set to 150, the number of starting nodes is 1000, and the algorithm runs for 20 complete iterations before every single neuron is added.

Other parameters are listed as follows:
\begin{enumerate}
\item 1)	The value of $top$ is found by studying the variance among the count of different individual domain states. We find that setting this to 22\% yields the best results.

\item 2)	We set $fix$ to be 30\% to simulate 1 in every 3 commands as strictly set by user regardless of MDP output. This has been set arbitrarily to enforce user authority.
\end{enumerate}

\section{Results}
\label{validation}

\begin{figure}
\centering
\includegraphics[width=0.45\textwidth]{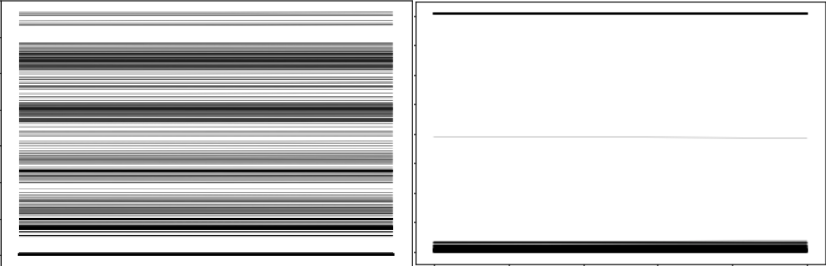}
\caption{Cluster Formations on the train data-set. Left: 263 domain state clusters obtained using GNG. Right: 3 device mode clusters obtained for Appliance 2 using GNG.} \label{cluster}
\end{figure}

We obtain 263, 242 and 240 cluster centres which make up the domain states, and, on an average, three modes per device, through Growing Neural Gas. We use 2,000,000 examples as our training data-se and. 1,000,000 examples as the test set. For validation, we divide the test set into 1,000 time slots, where each slot is essentially 1,000 consecutive readings. For each time slot, the clashes are calculated for SHD + SLD states. We also show LD state clashes along with Total Power Consumption averaged over each time slot. All the experiments for the data of each house have been conducted independently.

The results for clustering are shown in \ref{cluster}, which corresponds to graph outputs organized to show clusters for the GNG clustering algorithm. The left half of the image depicts 263 domain state clusters obtained in our experiments, and the right half of the image depicts three device mode clusters for one single appliance.

\begin{figure}
    \begin{subfigure}[b]{0.45\textwidth}
        \includegraphics[width=\textwidth]{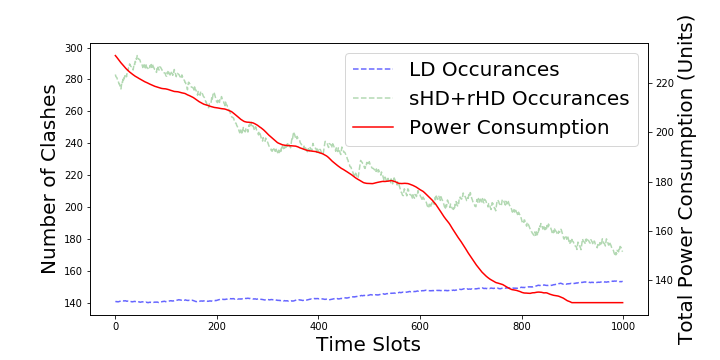}
        \caption{for smart home A}
        \label{fig:a}
    \end{subfigure}
    \begin{subfigure}[b]{0.45\textwidth}
        \includegraphics[width=\textwidth]{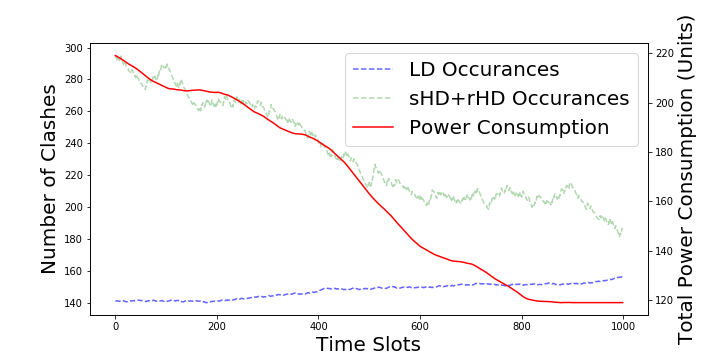}
        \caption{for smart home B}
        \label{fig:b}
    \end{subfigure}
    
    \begin{subfigure}[c]{0.45\textwidth}
        
        \includegraphics[width=\textwidth]{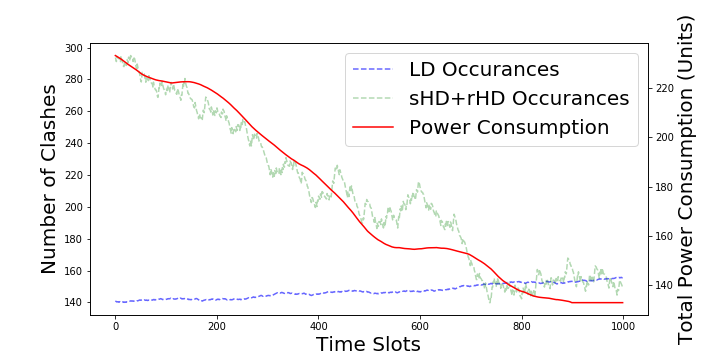}
        \caption{for smart home C}
        \label{fig:c}
        \end{subfigure}

    \caption{Resulting Clash Rates for LD, rLHD(randomLHD) Occurrences and corresponding fall in Total Power Consumption.}
    \label{result}
\end{figure}

From Figure \ref{result}, we infer the following information :

\begin{enumerate}
\item Reduction in total Clash Rate, since clashes for LD states, is more or less constant, and clashes in SHD + SLD occurrences reduce over time along with the total clashes, confirming that MDP is learning user behavior by outputting a greater number of states which the user would ``like'' or prefer.

\item MDP is not only learning user behavior but is also being specific about learning SHD states. Thus, the system chooses not to alter the usual user requirements.

\item When the MDP is beginning to learn the behavior, it is natural to have high number of clashes. However, over time when the MDP is intelligent enough to predict states, it outputs a “better” state than the current LD state, thereby creating the required scope for improvement regarding power consumption.

\item Total Power Consumption (in units) reduces over time, and this reduction is in a nearly linear fashion, in line with the decline in clash rate for SHD + SLD occurrences. The average power consumption is around 200 points on power consumption scale without the intelligent planner. In all the cases, we achieve convergence before 800 time-slots. This inference is logically consistent with our observation that the MDP is giving the user ’enough’ time to exhibit their usage habits and hence, adapts accordingly. 

In figure \ref{result}, an exciting behavior is observed between the LD and SHD + SLD occurrences in the plots for the houses. We elaborate on the same as follows: \newline

\item Around the time slot just before convergence, the plots in figure \ref{result} contrast to our theory that with a reduction in SHD + SLD clash occurrences, LD state clashes would compensate and increase. Figure \ref{fig:c} provides evidence at time slot 760 in which LD states increase linearly, whereas there are abrupt changes in SHD + SLD occurrences. This behavior, however, further bolsters the point that our model is adjusting to determine precisely which LD states need to be ignored and which SHD + SLD states are to be preferred.

\end{enumerate}

\begin{table}
 \caption{Comparison between different methods.}
 \begin{center}
     \begin{tabular}{|m{0.25\textwidth}|m{0.07\textwidth}|m{0.07\textwidth}|}
        \hline
         \bf{Method} & \bf{Energy Saved (\%)} & \bf{Considers User Behaviour} \\
         \hline
         Autonomous Appliance Scheduling for Household
         Energy Management \cite{12} & 10.92 & No \\
        \hline
        Intelligent Household LED Lighting System
        Considering Energy Efficiency and User Satisfaction \cite{13} & 21.9 & Yes \\
         \hline
        Making Smart Home Smarter: Optimizing Energy Consumption with Human in the Loop & 30 & Yes \\
         \hline
     \end{tabular}
 \end{center}
\label{comparisons}
\end{table}

The results show that considering user behavior while trying to reduce power consumption is a significant factor because the user must agree with the system when it comes to operating devices around him. In Table \ref{comparisons}, we show how the adaptive configuration of smart devices is quantitatively better than the other existing techniques. The mentioned methods are the closest to our problem statement, however, not exact. As explained in Section \ref{background}, Byun et. al. do not perceive user behavior to the extent as our method does, while Adika et. al. do not take user preference into account at all \cite{12, 13}. Their result shows around 22\% energy saving as opposed to a much better 30\% energy saving achieved by our method. Hence, our work justifies which states are ’compromised’ to obtain energy efficiency, and performs better at this task while considering user device setting preferences against other methods.

\section{Conclusion}

This paper presents an approach to adaptively configure the smart IoT devices through the effective incorporation of probabilistic user behavior(s). It takes into account the variance of devices regarding their application and more importantly, the user preference. However, it looks at the data only through power consumption perspective. Simulation of user actuation is intentionally kept stochastic to showcase that our system is resilient enough for real-life scenarios. The system uses each type of information available in the data-set, albeit with certain assumptions which at no point violate the requirement of stochasticity. Moreover, the system is made robust via the use of multilevel clustering. LD/HD states and the MDP algorithm help maintain an effective trade-off with user-preferred states and states which consume less power.

Even though current results very well substantiate the effectiveness of our system, we intend to extend our experiments to accommodate environments where even the clustered domain states are exponential. This problem can be tackled via enhanced state space approximation techniques and reduced computational requirements using factored MDPs. Future work can include formulating novel reward functions, which are a function of time as well, unlike the current static function since reward functions may be a function of time or the frequency of the number of shifts between device modes, etc. Also, we intend to search for a way to cluster devices based on their application in an automated manner to strengthen our automatic intelligent planner further.

\bibliographystyle{unsrt}  
\bibliography{bibliography_file}

\end{document}